\pdfoutput=1

\documentclass[11pt]{article}
\usepackage[whole]{bxcjkjatype}

\usepackage[]{naacl2021}

\usepackage{times}
\usepackage{latexsym}

\usepackage[T1]{fontenc}

\usepackage[utf8]{inputenc}

\usepackage{microtype}

\usepackage{multirow}
\usepackage{booktabs}
\usepackage{bm}
\usepackage{graphicx}
\usepackage{array}
\usepackage{CJKutf8}
\usepackage{comment}

\title{Sentence Concatenation Approach to Data Augmentation \\for Neural Machine Translation}

\author{Seiichiro Kondo, Kengo Hotate\thanks{~~Current affiliation: Recruit Co., Ltd.}, Tosho Hirasawa,\\
{\bf Masahiro Kaneko}\thanks{~~Current affiliation: Tokyo Institute of Technology} \and {\bf Mamoru Komachi}\\
Tokyo Metropolitan University \\
\texttt{kondo-seiichiro@ed.tmu.ac.jp, }\texttt{kengo\_hotate@r.recruit.co.jp}\\ \texttt{hirasawa-tosho@ed.tmu.ac.jp, } \texttt{masahiro.kaneko@nlp.c.titech.ac.jp}\\
\texttt{komachi@tmu.ac.jp}}

\begin{document}
\maketitle
\begin{abstract}
Neural machine translation (NMT) has recently gained widespread attention because of its high translation accuracy.
However, it shows poor performance in the translation of long sentences, which is a major issue in low-resource languages.
It is assumed that this issue is caused by insufficient number of long sentences in the training data.
Therefore, this study proposes a simple data augmentation method to handle long sentences.
In this method, we use only the given parallel corpora as the training data and generate long sentences by concatenating two sentences.
Based on the experimental results, we confirm improvements in long sentence translation by the proposed data augmentation method, despite its simplicity.
Moreover, the translation quality is further improved by the proposed method, when combined with back-translation.
\end{abstract}

\section{Introduction}

Neural machine translation (NMT) can be used to achieve high translation quality.
However, it has certain drawbacks, such as the degradation in the translation quality for long sentences.
\citet{koehn-knowles-2017-six} reported that the translation quality of NMT is superior to that of statistical machine translation (SMT) for input sentences within a certain length.
However, they also stated that when the sentence length exceeds a particular value, the quality of NMT becomes inferior to that of SMT, and the greater the sentence length, the lower the translation quality.

Additionally, they presented the correlation between the size of the training data and the translation quality \citep{koehn-knowles-2017-six}. 
In other words, the less training data we have, the lower will be the accuracy of the translation.
This issue is prevalent in low-resource languages. 
Therefore, various data augmentation methods for low-resource parallel corpora have been studied.
For instance, the generation of pseudo data was proposed by back-translating the monolingual corpora or paraphrasing the parallel corpora as additional training data \citep{wang-etal-2018-switchout, sennrich-etal-2016-improving, li-etal-2019-understanding}.

\begin{figure*}[h]
\centering
\includegraphics[width=16cm]{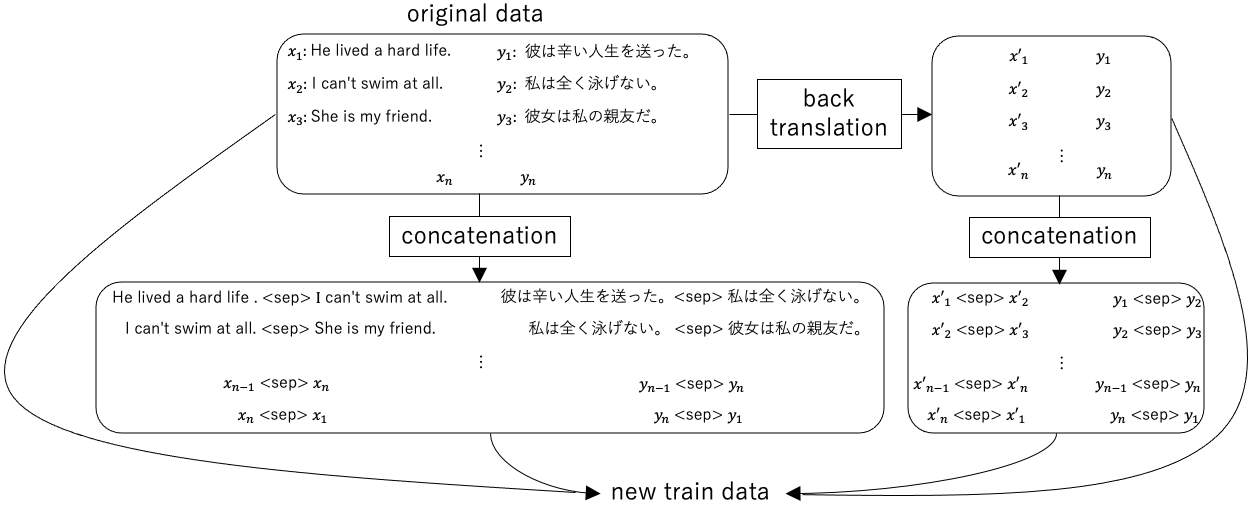}
\caption{Proposed method: Augmentation of data by combining the back-translation and the concatenation of two sentences. 
During concatenation, each sentence is randomly sampled, so that they do not have context overlap with each other.}
\label{fig:concat_aug}
\end{figure*}

Hence, this study proposes a data augmentation method that can be effective in long sentence translations.
The proposed method is illustrated in Figure \ref{fig:concat_aug}.
Long sentences were obtained by concatenating two sentences at random and adding them to the original data.
The translation quality is expected to be improved by this method because the low quality of translation of long sentences was caused by insufficient number of long sentences in the training data, which reduces this concern in the proposed method.

This study presents an improved BLEU score and higher quality in long sentence translations on English--Japanese corpus.
Moreover, the BLEU score further increases by incorporating back-translation.
In addition, human evaluation shows that fluency is increased more than adequacy.

In summary, the main contributions of this paper are as follows:
\begin{itemize}
\item We propose a simple yet effective data augmentation method, involving sentence concatenation, for long sentence translation.
\item We show that the translation quality can be further improved by combining back-translation and sentence concatenation.
\end{itemize}

\section{Related Works}

NMT exhibits a significant decrease in the translation quality for very long sentences.
\citet{koehn-knowles-2017-six} analyzed the correlation between the translation quality and the sentence length by comparing NMT with SMT.
They showed that the overall quality of NMT is better than that of SMT but that SMT outperforms NMT on sentences of 60 words and longer.
They stated that this degradation in quality was caused by the short length of the translations.
Additionally, \citet{neishi-yoshinaga-2019-relation} propose to use the relative position information instead of the absolute position information to mitigate the performance drop of NMT models for long sentences.
They conducted an analysis of the translation quality and sentence length on length-controlled English--to--Japanese parallel data and showed that the absolute positional information sharply drops the BLEU score of the transformer model \citep{10.5555/3295222.3295349} in translating sentences that are longer than those in the training data.

Several data augmentation methods have been proposed for NMT, such as back-translation, which involves translating the target-side monolingual data to create a pseudo dataset \citep{sennrich-etal-2016-improving}.
In their method, the back-translation model is first learned by using parallel corpora from the target-side to the source-side.
Once converged, this model generates pseudo data by translating the target-side monolingual corpora to the source-side language.
A translation model is then trained using both the pseudo-parallel and original-parallel data.
\citet{li-etal-2019-understanding} analyzed multiple data augmentation methods.
In their experiments, they applied self-training and back-translation.
In self-training, they fixed the source-side and used a forward translation model to generate the target-side, and in back-translation, they fixed the target-side and used a backward translation model to generate the source-side.
It was observed that these methods can effectively improve the translation accuracy for infrequent tokens.
These methods can be used with the sentence concatenation method proposed in this study.

In multi-source neural machine translation, \citet{DBLP:journals/corr/DabreCK17} proposed concatenating source sentences in different languages corresponding to a target sentence in training.
However, they did not aim to improve the translation accuracy of long sentences.
Our method concatenates two source sentences in the same language at random.

\begin{table*}[h]
\centering
\small
\begin{tabular}{lccccccccc}
\toprule
length & all & 1 -- 10 & 11 -- 20 & 21 -- 30 & 31 -- 40 & 41 -- 50 & 51 -- 60 & 61 -- 70 & 71 -- \\
\midrule
sentences              & 1,812 &   73 &  529 &  600 &  341 &  164 &   74 &   18 &   13 \\
\midrule
vanilla (400K)         &      26.5 &      22.9 &      23.0 &      26.2 &      27.1 &      29.6 &      28.5 &      28.8 &      23.6 \\
\ + concat (+ 400K)      &      26.6 &      21.4 &      23.3 &      25.7 &      27.5 &      29.5 &      28.7 &      28.2 &      29.0 \\
\ + ST (+ 400K)          &      28.2 &      23.9 &      24.8 &      27.4 &      28.6 &      31.4 &      31.4 &      29.6 &      27.6 \\
\ + BT (+ 400K)          &      28.8 &      24.3 &      25.5 &      28.3 &      29.5 &      31.6 &      30.6 &      28.7 &      28.7 \\
\ + BT + concat (+ 1.2M) & $\bm{29.4}$ & $\bm{25.4}$ & $\bm{25.6}$ & $\bm{28.6}$ & $\bm{30.1}$ & $\bm{33.1}$ & $\bm{31.5}$ & $\bm{29.9}$ & $\bm{30.1}$ \\
\bottomrule
\end{tabular}
\caption{BLEU scores for each sentence length breakdown on the test data set: ``vanilla + BT + concat'' consists of data from vanilla, BT, and concatenation of both.}
\label{tab400K}
\end{table*}

\section{Data Augmentation by Sentence Concatenation}
The proposed method augments the parallel data by back-translation and concatenation.
A schematic overview of the proposed method is shown in Figure \ref{fig:concat_aug}.

First, we back-translate the target-side of the parallel corpus \citep{li-etal-2019-understanding, sennrich-etal-2016-improving} to create pseudo data as additional training data.
Note that we do not use external data in back-translation, and the diversity of target sentences does not change.

Then, we randomly select two sentences exclusively in the original or pseudo data and concatenate them to create another training data.
Technically, we concatenate two source sentences and insert a special token, ``<sep>,'' between them. 
Corresponding target sentences are concatenated in the same way.
Afterwards, we remove the sentences consisting of less than 25 words from the pseudo data.

Finally, we obtain an augmented training data comprising original, pseudo, and concatenated sentences, which has the quadruple data size of the original training data.

We train our models on both single and concatenated sentences first because models can learn to translate single sentences.
We also expect models to acquire a better absolute position encoding to translate long sentences in the better quality without generating a special token (i.e., <sep>) contained in concatenated sentences in the inference process.

During the testing process, a single sentence is fed as the input, even though the training data contains concatenated sentences.\footnote{We also conducted an experiment with two sentences as input during the test, but the BLEU score was worse than the proposed method.}

\begin{figure}[t]
\centering
\includegraphics[width=8cm]{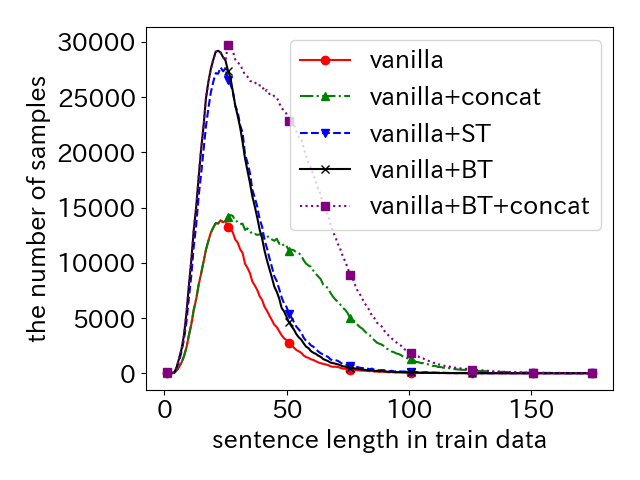}
\caption{Distribution of each data set.}
\label{fig:length_dist}
\end{figure}

\section{Experiments}

\subsection{Models}
To investigate the effectiveness of the proposed method when combined with previous data augmentation methods, five types of training data were prepared from the original training data.

Figure \ref{fig:length_dist} shows the number of training data used in this experient.
Note that the total number of sentences in ``vanilla + concat,'' ``vanilla + ST'' and ``vanilla + BT'' are nearly equal.
In the source language, the average sentence length of ``vanilla'' is 30.39, and that of ``vanilla+concat'' is 46.18.

We train the forward translation models using the training data and compare the BLEU scores obtained for the output of the test data.
\paragraph{vanilla.} 
Original data.
\paragraph{vanilla + concat.} 
Original data and augmented data by sentence concatenation.
Sentences with length of less than 25 words after concatenation were removed to improve the translation quality of long sentences.
\paragraph{vanilla + ST.} 
Original data and augmented data by self-training.
\paragraph{vanilla + BT.} 
Original data and augmented data by back-translation.
\paragraph{vanilla + BT + concat.} 
The composite data of the original data, the back-translated data, and their sentence concatenation.\footnote{The results of the experiment showed that the score of ``vanilla + BT'' was higher than that of ``vanilla + ST.''
Therefore, in this study, the proposed method was combined only with ``vanilla + BT.''}

\begin{table}[t]
\centering
\small
\begin{tabular}{lcccccc}
\toprule
       &  \multicolumn{3}{c}{adequacy} & \multicolumn{3}{c}{fluency}\\ \cmidrule(lr){2-4}\cmidrule(lr){5-7}
length & win & tie & lose & win & tie & lose \\
\toprule
\ \ 1 -- 10 & $\bm{4}$ &         5 &         3 &  $\bm{3}$ &         7 &         2 \\
11 -- 20   &        20 &        39 & $\bm{29}$ & $\bm{21}$ &        47 &        20 \\
21 -- 30   & $\bm{34}$ &        35 &        31 & $\bm{33}$ &        42 &        25 \\
31 -- 40   & $\bm{23}$ &        21 &        13 & $\bm{17}$ &        24 &        16 \\
41 -- 50   &        10 &         6 & $\bm{11}$ &         6 &        10 & $\bm{11}$ \\
51 --      &  $\bm{6}$ &         5 &  $\bm{6}$ &  $\bm{7}$ &         6 &         4 \\
\midrule
overall    & $\bm{97}$ &       111 &        93 & $\bm{87}$ &       136 &        78 \\

\bottomrule
\end{tabular}
\caption{Human evaluation: Pairwise comparison of ``vanilla + BT'' and ``vanilla + BT + concat.'' 
``win'' denotes the sentence generated by our proposed method, ``vanilla + BT + concat,'' is superior to that of ``vanilla + BT,'' and ``lose'' denotes the opposite of ``win.''}
\label{Human_eval}
\end{table}

\newcolumntype{C}[1]{>{\hfil}m{#1}<{\hfil}}
\newcolumntype{L}[1]{m{#1}<{\hfil}}
\newcolumntype{R}[1]{>{\hfil}m{#1}}

\begin{figure}[t]
\centering
\includegraphics[width=7.5cm]{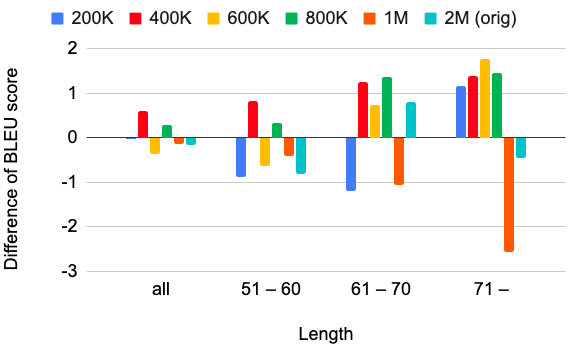}
\caption{Effectiveness of the proposed method for each data size by sentence length: Vertical axis represents BLEU score of ``vanilla + concat + BT'' minus BLEU score of ``vanilla + BT.''}
\label{fig:each_data_size}
\end{figure}

\begin{table}[h]
\centering
\small
\begin{tabular}{rrcc}
\toprule
\multirow{2}{*}{length} & \multirow{2}{*}{sentences} & \multicolumn{1}{l}{vanilla} & \multicolumn{1}{l}{vanilla} \\
&& \multicolumn{1}{l}{\ + BT } & \multicolumn{1}{l}{\ + BT + concat} \\
\toprule
       all & 999,998 &        22.1 & $\bm{22.2}$ \\
\midrule
   1 -- 10 &  22,725 &        18.2 & $\bm{18.3}$ \\
  11 -- 20 & 232,829 & $\bm{17.9}$ & $\bm{17.9}$ \\
  21 -- 30 & 329,597 &        20.1 & $\bm{20.2}$ \\
  31 -- 40 & 219,845 &        22.1 & $\bm{22.3}$ \\
  41 -- 50 & 109,528 &        23.2 & $\bm{23.4}$ \\
  51 -- 60 &  47,851 &        24.3 & $\bm{24.4}$ \\
  61 -- 70 &  20,526 &        24.6 & $\bm{24.8}$ \\
 71 -- 100 &  15,557 &        25.1 & $\bm{25.4}$ \\
101 -- 200 &   1,540 &        20.1 & $\bm{22.3}$ \\
\bottomrule
\end{tabular}
\caption{BLEU scores for each sentence length breakdown on the pseudo test data set: pseudo test data consists of 1M sentences from the training data that were not used for training.}
\label{tab_test_1M}
\end{table}

\begin{figure}[t]
\centering
\includegraphics[width=7.5cm]{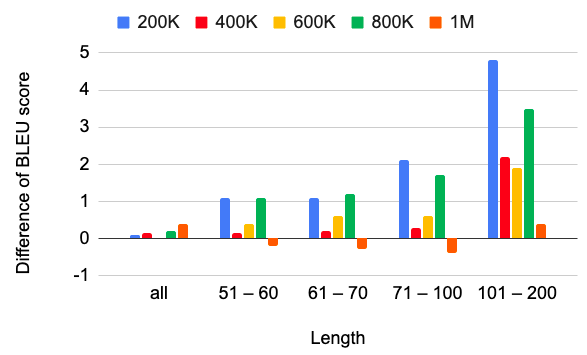}
\caption{Effectiveness of the proposed method for each data size by sentence length in 1M pseudo test set.}
\label{fig:each_data_size_1M}
\end{figure}

\begin{table*}[h]
\small
\centering
\begin{tabular}{C{1.0cm}L{14.0cm}}
\toprule
src             & Myanma is behind in market economization together with Laos, Canbodia, Vietnam, and the GDP per one person is the lowest in the 4 countries, and it remains \$ 180, but Myanma is thought to remarkably develop if political problems are solved, because flatland occupies $7 \times10\%$ of the land and natural resources are rich, and because personnel expenses are extremely cheap.\\
\midrule

tgt             & ミャンマーは自国とともに後発ＡＳＥＡＮ４カ国といわれるラオス，カンボディア，ベトナムと比較しても市場経済化が遅れ，一人あたりのＧＤＰは最低で１８０ドルにとどまっているが，平地が７割で天然資源もあり，人件費が極端に安価なので，政治的問題が解決されれば著しく発展すると見られる。 \\
\midrule

vanilla         & ミャマは，陸上と自然資源の７割を占めるため，平地は土地と自然資源の７割を占めるので，人件費が極端に安く，４か国で１人当たりＧＤＰが最低である。\\
\midrule

vanilla\par +concat &
ミャンマーは\textbf{ラオス，カンボジア，ベトナムと共に市場経済化に遅れ}，４国ではＧＤＰが１人あたり最低であるが，国土の７割を占める平坦な土地と自然資源が豊富で人件費が極端に安く\textbf{政上の問題が解決されれば，顕著に発展すると考えられる。}\\
\bottomrule
\end{tabular}
\caption{An example of the effectiveness of the proposed method.}
\label{ex1}
\end{table*}

\subsection{Setup}
We used ASPEC\footnote{\href{http://lotus.kuee.kyoto-u.ac.jp/WAT/WAT2017/snmt/index.html}{http://lotus.kuee.kyoto-u.ac.jp/WAT/WAT2017/snmt/}} from WAT17 \citep{nakazawa-etal-2017-overview} to perform English-to-Japanese translation.
This dataset contains 2M sentences as training data, 1,790 as valid data and 1,812 as test data.
We also followed the official segmentation using SentencePiece \citep{kudo-richardson-2018-sentencepiece} with a vocabulary size of 16,384.
A total of 400K sentences were randomly extracted from the original training data and selected as the training data to be used in this experiment.
Regarding self-training and back-translation models, we used only the training corpus, following \citet{li-etal-2019-understanding}.

The transformer models from Fairseq were used in the experiment \cite{ott-etal-2019-fairseq}\footnote{\href{https://github.com/pytorch/fairseq}{https://github.com/pytorch/fairseq}}.
Adam was set as the optimizer with a dropout of 0.3, a maximum of 300,000 steps in the training process, and a total batch size of approximately 65,536 tokens per step.
The same architecture was also used to train the self-training and the back-translation models.

The BLEU score \citep{papineni-etal-2002-bleu} was used for automatic evaluation.
We computed the average of the BLEU scores of three runs with different seeds.
Human evaluation was also conducted.
For three native Japanese evaluators, 100 sentences were randomly selected from the test set per evaluator. 
They performed pairwise evaluation between ``vanilla + BT'' and ``vanilla + BT + concat'' from two perspectives: adequacy and fluency.

\subsection{Results}
\paragraph{Automatic evaluation.}
The result of this experiment is presented in Table \ref{tab400K}.
It describes the BLEU scores measured for each test data classified by the sentence length.

The BLEU score of ``vanilla + concat'' is more stable when applied for translation with sentence lengths of longer than 51 words, which are the majority of data augmented by the sentence concatenation, although the score for the sentences classified as 61--70, is slightly lower than that of ``vanilla.''
Conversely, the quality of the translation of short sentences is greatly reduced.

Additionally, the overall score of ``vanilla + BT + concat'' is higher than that of ``vanilla + BT'' by 0.6.
In particular, the score of the sentence lengths of longer than 41 is significantly improved, which indicates that the proposed method is more effective for long sentence translation.
In Addition, the score of ``vanilla + BT + concat'' is much higher than that of ``vanilla + concat.''
Consequently, it is shown that the back-translation and concatenation are independent factors that improve the accuracy of the translation. 

\paragraph{Human evaluation.}
Table \ref{Human_eval} presents the results of human evaluation.
We observed that the output of the proposed method improved or were comparable under almost all conditions except for ``11--20'' on adequacy and ``41--50'' on fluency.
The proposed method added the sentences whose length is more than 25 words and is effective in improving the translation of such sentences.

\begin{table*}[h]
\small
\centering
\begin{tabular}{C{1.0cm}L{14.0cm}}
\toprule
src             & Results of the analysis shows high accuracy properties, such as the reproducibility of relative standard deviation 0.3\textasciitilde0.9\% varified by repetitive analyses of ten times, the clibration curves with correlation coefficient of 1 verified by tests of standard materials in using six kinds of acetonitrile dilute solutions, and the formaldehyde detection limit of 0.0018$\mu$g/mL.\\
\midrule

tgt             & 結果は，相対標準偏差０．３〜０．９％の再現性（１０回の繰返し分析），相関係数１の検量線（６種類のアセトニトリル希釈溶液による標準資料の検定），０．００１８μｇ／ｍＬのホルムアルデヒド検出限界，など高い精度を得た。\\
\midrule

vanilla\par +BT         & ６種のアセトニトリル希薄溶液を用いた標準物質の試験及びホルムアルデヒド検出限界は０．００１８μｇ／ｍＬであった 。\\
\midrule

vanilla\par +BT\par +concat &
分析の結果は\textbf{１０回の繰り返し解析で相対標準偏差０．３〜０．９％の再現性}，６種のアセトニトリル希薄溶液を用いた標準物質の試験により検証された\textbf{１の相関係数を持つクライテリア曲線}，及び０．００１８μｇ／ｍＬのホルムアルデヒド検出限界など高い精度を示した。\\

\bottomrule
\end{tabular}
\caption{An example where the proposed method worked well.}
\label{ex3}
\end{table*}

\begin{table*}[h]
\small
\centering
\begin{tabular}{C{1.0cm}L{14.0cm}}
\toprule
src             & These seemed to be noticeable complications in case of extracorporeal circulation for umbilical hernia repair.\\
\midrule

tgt             & さい帯ヘルニア修復術における体外循環の合併症として注目すべきと思われた 。\\
\midrule

vanilla       & さい帯ヘルニア修復術における体外循環の合併症として注目すべきと思われた 。\\
\midrule

vanilla\par +concat & 以上の所見より，\underline{さい帯ヘルニア}に対する体外循環では，\underline{特に合併症として}\underline{特に合併症として}，特に，\underline{さい帯ヘルニア}では体外循環がより注意を要すると考えられた。\\
\bottomrule
\end{tabular}
\caption{An example where the proposed method may have caused errors.}
\label{ex4}
\end{table*}

\subsection{Discussion}
\paragraph{Test set.}
Figure \ref{fig:each_data_size} depicts the breakdown in the difference between the BLEU scores of the proposed method for each training data size per sentence length.
Notably, for sentences with 51 words or longer, the translation quality improves when the size of data is between 400K and 800K.
However, the translation quality degrades when there are more than 1M sentences.
The proposed method is not suitable when a large amount of training data is available.

In the human evaluation, we observe that the proposed method is more effective in terms of fluency than adequacy.
It is assumed that the translation model can handle absolute positional encoding for long sentences by the proposed method.

\paragraph{Pseudo test set.}
In this experiment, the number of bilingual sentences in the test set was small, especially in long sentences.
For this reason, additional experiments were carried out to confirm the validity of the results.
For evaluation, we extracted 1M sentences from the training data that were not used for training and used them as the pseudo test data.
Table \ref{tab_test_1M} shows the average of the BLEU scores for the three runs with 400K training data with different seeds. 
Note that the overall BLEU score is, however, lower than when using the test data, but this is probably because the quality of the training data is lower than that of the test data.

By comparing the results of ``vanilla + BT'' and that of the proposed method, the proposed method was shown to have a slightly better overall score. 
Examining the scores by sentence length, there was a significant increase in scores for longer sentences, especially for ``101 -- 200'' sentences. 
It indicates that the proposed method is effective in improving the translation accuracy of long sentences.

Also, a comparison similar to the one using the test set was conducted using this 1M pseudo test data. 
The results are shown in Figure \ref{fig:each_data_size_1M}.
In this setting, it is more evident that for sentences with a sentence length of 51 words or more, the translation accuracy improves when the data size is 800K or less and decreases when the data size exceeds 1M. 

\subsection{Case Study}

Tables \ref{ex1} and \ref{ex3} show the cases in which the proposed method worked effectively in this experiment, whereas Table \ref{ex4} shows the cases in which the translation quality deteriorated.

The example in Table \ref{ex1} shows that the sentence output by ``vanilla'' is shorter than expected, which indicates that necessary information for translation is missing.
Conversely, the output of ``vanilla + concat'' is a longer sentence, which reduces the missing information.

The example in Table \ref{ex3} shows an example of improved translation by using the proposed method.
Similar to the previous example, ``vanilla + BT'' completely loses the information in the first half of the sentence, while ``vanilla + BT + concat'' produces a translation that includes the information of the entire sentence.

However, as shown in the example in Table \ref{ex4}, there were cases where the output of the model trained including concatenated data showed repetitive outputs that were not seen in the output of the model trained on the original data.
This type of output occurs more frequently in the case of short sentences.
This suggests that the ability to output long sentences may lead to unnatural repetition of the output because of the attempt to generate long sentences.

\section{Conclusion}

This study proposes a data augmentation method to improve the translation quality of long sentences.
The experimental results confirmed that the data augmentation method is straightforward but useful, especially for the translation of very long sentences.
However, the quality of the translation of short sentences is reduced.

In the future, we would like to develop a method that works well when there is a large amount of available parallel data.
Moreover, since the adequacy of the translation of short sentences is considerably low in the proposed method, we would like to compensate for this weakness by considering the reconstruction loss \citep{tu2017neural}. 
Also, it would be interesting to explore the use of interpolation of hidden space for data augmentation
considering long sentences \citep{chen-etal-2020-mixtext}.



\section*{Acknowledgements}
We would like to thank Aomi Koyama, Tomoshige Kiyuna and Hiroto Tamura for their evaluation.
We would also like to thank the anonymous reviewers for their valuable feedback.

\bibliography{anthology,custom}
\bibliographystyle{acl_natbib}




\end{document}